# TECM*: A Data-Driven Assessment to Reinforcement Learning Methods and Application to Heparin Treatment Strategy for Surgical Sepsis

Jiang Liu[†a], Yujie Li[†b], Chan Zhou[a], Yihao Xie[a], Qilong Sun[a], Xin Shu[b], Peiwei Li[a], Chunyong Yang[b], Yiziting Zhu[b], Jiaqi Zhu[c], Yuwen Chen[*a], Bo An[*d], Hao Wu[*e], Bin Yi[*b]

*Abstract—Objective:* **Sepsis is a life-threatening condition caused by severe infection leading to acute organ dysfunction. This study proposes a data-driven metric and a continuous reward function to optimize personalized heparin therapy in surgical sepsis patients.** ***Methods:*** **Data from the MIMIC-IV v1.0 and eICU v2.0 databases were used for model development and evaluation. The training cohort consisted of abdominal surgery patients receiving unfractionated heparin (UFH) after postoperative sepsis onset. We introduce a new RL-based framework: converting the discrete SOFA score to a continuous cxSOFA for more nuanced state and reward functions; Second, defining "good" or "bad" strategies based on cxSOFA by a stepwise manner; Third, proposing a Treatment Effect Comparison Matrix (TECM), analogous to a confusion matrix for classification tasks, to evaluate the treatment strategies. We applied different RL algorithms, Q-Learning, DQN, DDQN, BCQ and CQL to optimize the treatment and comprehensively evaluated the framework.** ***Results:*** **All AI strategies reduce mortality and average hospital stay significantly. Among the AI strategies, the cxSOFA-CQL model achieved the best performance, reducing mortality from 1.83% to 0.74% with the average hospital stay from 11.11 to 9.42 days, the performances improved 59.56% and 15.21%, respectively. TECM demonstrated consistent outcomes across models, highlighting robustness.** ***Conclusion:*** **The proposed RL framework enables interpretable and robust optimization of heparin therapy in surgical sepsis. Continuous cxSOFA scoring and TECM-based evaluation provide nuanced treatment assessment, showing promise for improving clinical outcomes and decision-support reliability.**

This work was partially supported by National Science Foundations of China (No. 62371438, No. 82470655, and No. 82100658), Science and Chongqing Talents Project (No. CQYC202103080), Chongqing Municipal Natural Science Foundation (No.CSTB2024 NSCQ-LZX0037 & No. CSTB2023 NSCQ-ZDJ0005), Youth Top Talent Project of Chongqing Municipal Health Commission (No.YXQN202434).

Chongqing Institute of Green and Intelligent Technology, CAS

Jiang Liu (Ph.D., liujiang@cigit.ac.cn), Chan Zhou (M.Eng, zhouchan@cigit.ac.cn), Yihao Xie (M.Eng, yugumarck@gmail.com), Qilong Sun (Ph.D., sunqilong@cigit.ac.cn,), Peiwei Li (M.Eng, lipeiwei25@mails.ucas.ac.cn), Yuwen Chen (Ph.D. , chenyuwen@cigit.ac.cn*).

Department of Anesthesiology, Southwest Hospital, Third Military Medical University

Yujie Li (M.D., lyj09c@tmmu.edu.cn), Xin Shu (M.M., xinshuxin31@163.com), Chunyong Yang (M.D., bnycy@163.com), Yiziting Zhu (M.M., zhuyizyzt@tmmu.edu.cn), Bin Yi (M.D., yibin1974@tmmu.edu.cn*).

Institute of Software, CAS

Jiaqi Zhu (Ph.D., zhujq@ios.ac.cn)

College of Computing and Data Science, Nanyang Technological University

Bo An* (Ph.D., boan@ntu.edu.sg)

First Affiliated Hospital of Chongqing Medical University

Hao Wu* (M.D., ewuhao@163.com)

## I. INTRODUCTION

Sepsis is a severe infection that leads to life threatening acute organ dysfunction, with an estimated millions of incident cases and deaths worldwide [1]. Despite reductions in early deaths, surgical sepsis rates are climbing, with at least one-third of patients developing chronic critical illness, which imposes a significant burden on public health [2]. According to international sepsis guidelines [3], fluid resuscitation, antibiotics, vasopressors, and other treatments are effective therapies, and the doses and timing should be adjusted based on dynamic measurements of disease progression. Decision-making in managing sepsis patients is high-stakes, complex, uncertain, and informed by rapidly changing, high-volume data inputs [4]. Therefore, the key to effective sepsis treatment lies in the precise, prompt, and efficient assessment of patient status and treatment response. However, the underlying mechanisms of sepsis remain elusive, and the heterogeneity among sepsis patients limits the effectiveness of widely used metrics like the Sequential Organ Failure Assessment (SOFA) score [5] and the National Early Warning Score (NEWS) [6], which are insufficient for drug discovery and the development of new therapies [7]. Recently, several machine learning-based assessment methods have been developed and validated [8, 9], offering hope for personalized sepsis treatment. Nevertheless, most of these models focus on the early warning of sepsis based on non-time-series variables, and few are designed for the dynamic evaluation of treatment response. Furthermore, models or methods evaluating responses to treatments such as fluid resuscitation or antibiotic therapy often include specific and non-routine features with a time lag, which also impedes drug discovery and the development of precisely personalized treatments [10, 11].

In the realm of optimizing sepsis treatment, the timing and dosing of antibiotics are the most investigated [11, 12], because the state and response to treatment are relatively clear, with a limited time window. However, sepsis treatment involves more than just antibiotics, especially for multi-organ protection. Heparin, first used in the treatment of sepsis in 1966, continues to be a subject of debate regarding its efficacy and safety in sepsis patients. Some studies suggest that heparin may reduce mortality at 28 days[13, 14], while others do not find significant effects [15, 16]. Previously, we showed that heparin treatment benefits surgical sepsis patients and proposed an optimized strategy based on RL models [17].

In terms of optimizing critical care decision-making, reinforcement learning (RL) stands out as a compelling solution for therapy timing and dosage [11, 12, 17], treatment selection, and optimization [18-20], and it aims to maximize long-term benefits for patients. Unlike traditional randomized controlled trials, RL provides precise treatment tailoring [21] and approximates optimal treatment strategies based on historical data without requiring prior knowledge of biological system models [22]. However, applying RL in medicine still presents challenges. First, medical scenarios demand more than sparse reward functions that only consider final treatment results [23]. Second, ethical constraints prohibit direct experimentation in patients, so properly evaluating AI strategies remains a significant challenge [24]. Although some progress [17, 23] has been made, several issues persist:

1） Reward Function Issue: In the area of RL, the state serves as the basis for an agent's decision-making, while the reward function gauges the effectiveness of its actions. Currently, most research on RL-based optimization treatments for sepsis relies on discrete states, such as SOFA score [23] or a limited number of clusters [19], which form the foundation for defining the reward function. However, the states of the septic patients are complicated and rapidly changing, which cannot be fully captured by discrete values alone. Thus, the design of reward functions for optimizing and assessing treatment strategies with continuous state space and action space remains largely unexplored.

2） Good/Bad Treatment Definition Issue: The assessment of AI strategies is predominantly conducted by experts. This approach poses challenges when no consensus is reached. Even though a study [17] provides an initial study of data-driven assessment, its definitions of "good" and "bad" practical physician treatments are based solely on the outcome of death or survival. If all patients in the datasets survive, differentiating between treatment strategies with varying stepwise improvements becomes challenging.

3） Incomplete Assessment Issue: The Q(s,a)-function in Q-learning[25], also known as the action-value function, is a mathematical representation that estimates the expected accumulative return or utility of taking action a in states. In non-medical applications such as natural language processing and autonomous driving, RL models focus on optimizing the best strategies, often due to efficiency and resource constraints, they may overlook the exploration of less effective strategies. In contrast, critical care medicine requires a comprehensive consideration of all strategies, including the less effective ones, to enhance treatment efficacy and safety. Consequently, focusing on optimal strategies in RL applications could limit a comprehensive strategy evaluation, potentially impacting the real-world effectiveness and safety of the models. This comprehensive approach is essential for evaluating AI strategies in medical settings, as it addresses the substantial challenge of integrating both "good" and "bad" physician practices with the full spectrum of AI strategies, from optimal to worst-case scenarios. Achieving a clear and understandable synthesis of these elements is vital for creating a thorough assessment of AI strategies in critical care medicine, e.g., sepsis, and acute kidney injury.

To address these issues, we propose a novel reinforcement learning framework with a continuous state representation and a robust policy evaluation matrix in this study. The key contributions of this study are as follows:

- we extend the SOFA to a continuous version, called the continuous extension of SOFA (cxSOFA for short), to address the reward function issue.
- For the good/bad treatment issue, we propose a data-driven definition based on the change rates of a patient's health status within a treatment trace.
- Additionally, we introduce comprehensive assessment metrics, denoted as TECM*, to tackle the issue of incomplete assessments. Furthermore, we will establish new training termination conditions for RL based on TECM*.

## II. Data and Methods

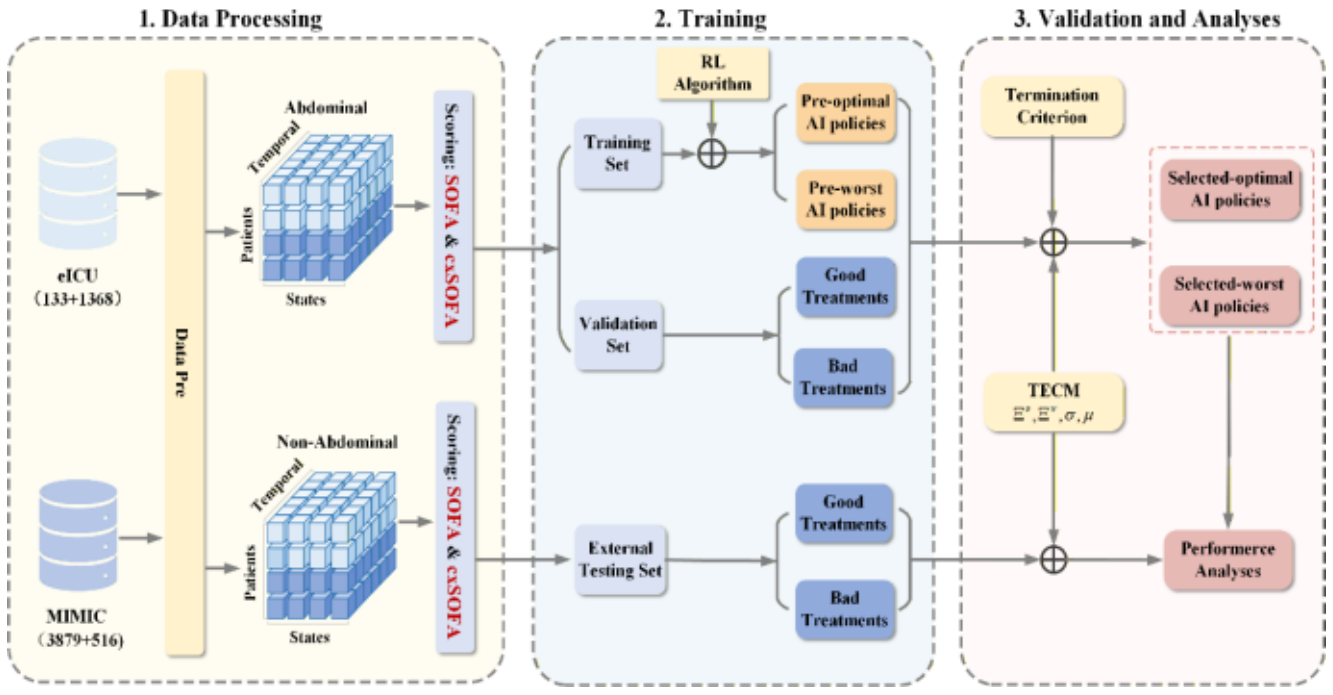

Fig. 1. Overview of the framework.

To obtain dynamic and personalized optimization of treatment strategy for surgical sepsis, we propose an RL-based framework with a continuous SOFA score as the reward function and a novel assessment metric based on the TECM. Herein, we evaluate the framework by optimizing the heparin treatment strategy for patients with surgical sepsis. As shown in Fig.1, the framework comprises data extraction and pre-processing, RL training, model validation based on the TECM, and analyses. First, treatment time-series data were extracted from two independent open-access datasets, eICU and MIMIC-IV. This included data from two distinct patient cohorts: (1) those with abdominal surgery, and (2) those with non-abdominal surgery. The raw data underwent preprocessing, including cleaning, nearest-neighbor imputation, and linear interpolation, followed by calculation of SOFA and cxSOFA scores for state representation.The time-series data from the abdominal surgery cohort were then partitioned into a training set (80%) and a validation set (20%). The data from the non-abdominal surgery cohort were designated as the external testing set. Subsequently, SOFA and cxSOFA scores were used to construct corresponding MDP models. Five RL algorithms using SOFA and cxSOFA scores respectively were trained on the training datasets. For each RL algorithm, the newly proposed η-TECM* termination criterion was used to select an appropriate $Q(s, a)$-function and identify the optimal and worst AI strategies. Finally, the best strategy among the optimal strategies from various algorithms was selected using TECM* and tested on the testing datasets. In the following sections, we present the details of key processes.

### *A. Datasets Preprocessing*

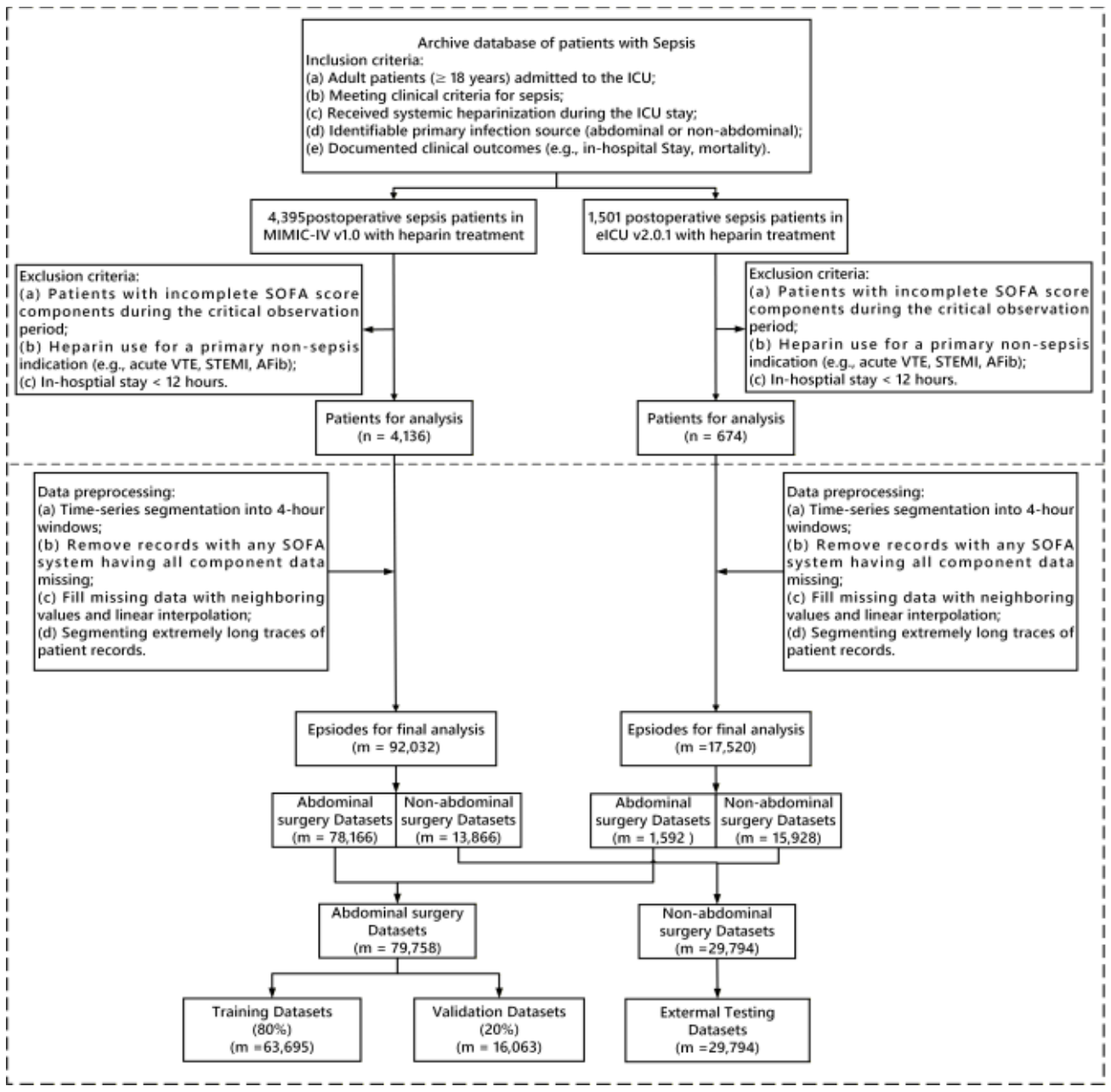

Fig. 2. The flowchart of patient enrollment and data preprocessing.

This study will employ multi-center data obtained from two open datasets: MIMIC-IV v1.0 [26] and eICU v2.0.1 [27]. The patient selection process is detailed in Fig.2. The cohort consisted of postoperative sepsis patients, including groups receiving varying doses of heparin treatment after their sepsis diagnosis. Relevant information, such as timestamps of admission and diagnosis, sequential SOFA scores, heparin administration records, and clinical outcomes, were extracted. A 4-hour window for evaluating heparin treatment was defined, consistent with prior studies [28-30].

After dividing the time windows, we found that there were many missing values in the datasets, which may have a significant impact on data analysis. To manage missing data inherent to these databases, a two-phase approach was employed. First, preliminary data imputation was conducted for values within defined temporal boundaries using last-value-carried-forward and linear interpolation techniques. Second, a stringent exclusion criterion was applied: any temporal window lacking documented heparin dosage or missing any primary physiologic variable necessary for SOFA score computation was discarded in its entirety. This process ensured the integrity of the data structure required for the MDP modeling framework [31]. After processing the incomplete data, the length of each patient's time series was calculated. Finally, extremely long trace of patient records will be segmented.

### *B. MDP Formulation*

We formulate the strategy optimization as an RL task. Specifically, we model our problem as a sequential decision task using a finite MDP framework, defined by a tuple $(S, A, R, T, \gamma)$. The elements of MDP are interpreted as follows:

The state space $S$ constitutes a finitely dimensional space. In the Q-Learning model, each state corresponds to the SOFA score at that time step, whereas in deep reinforcement learning models (DQN, DDQN, BCQ, CQL), each state is a tuple of the six component scores from the SOFA or cxSOFA score, corresponding to organ systems provided in Table I. Consequently, the state $s_t = (f_1, f_2, f_3, f_4, f_5, f_6)$ at any time step $t$ is instantiated as SOFA or cxSOFA-derived vector computed from the patient's real-time physiological parameters: a one-dimensional scalar for QL and a six-dimensional vector for deep reinforcement learning models.

TABLE I
VITAL SIGN AND MODEL STATES

| Symbol | Variable | Vital sign | Organ system |
|---|---|---|---|
| $f_1$ | $x_0$ | Oxygenation (no vent) | respiratory system |
| | $x_1$ | Oxygenation (vent) | |
| $f_2$ | $x_2$ | platelet | coagulation system |
| $f_3$ | $x_3$ | bilirubin | liver function |
| $f_4$ | $x_4$ | MBP | circulatory system |
| | $x_5$ | dopamine | |
| | $x_6$ | dobutamine | |
| | $x_7$ | epinephrine | |
| | $x_8$ | norepinephrine | |
| $f_5$ | $x_9$ | GCS | nervous system |
| $f_6$ | $x_{10}$ | creatinine | renal function |
| | $x_{11}$ | UO | |

The action space $A$ is a finite set defined by intervals of heparin dosage injections, as specified in Table Ⅱ. Specifically, the action at time step $t$ takes a value from the set $\{a_0, a_1, a_2, a_3, a_4\}$.

TABLE II
DOSING AND MODEL ACTION

| DOSING OF UFH (U/KG/H) | Action |
|---|---|
| 0 | $a_0$ |
| $(0, 1.38]$ | $a_1$ |
| $(1.38, 1.88]$ | $a_2$ |
| $(1.88, 3.5]$ | $a_3$ |
| $> 3.5$ | $a_4$ |

The transition probability $T$ resulting from an action is deterministic in our case, as the agent moves from one state to the next according to the order of samples presented in the training data.

Similar to the previous studies [17, 23], the reward function R at time t is determined jointly by the SOFA or cxSOFA scores at time t and at time t + 1, as illustrated in

$$r_t^{SOFA} = SOFA(s_t) - SOFA(s_{t+1}) \quad (1)$$

$$r_t^{cxSOFA} = cxSOFA(s_t) - cxSOFA(s_{t+1}) \quad (2)$$

for a treatment episode

$$ep = (s_0, a_0; s_1 . a_1; \dots ; s_{n-1}, a_{n-1})$$

where $s_t$ is the state of 6 dimensions. Through the SOFA or cxSOFA score, the lower the score, the better the patient's condition. Therefore, if the patient's condition improves, the score decreases, resulting in a positive reward $r_t > 0$. Conversely, if the patients' condition worsens, a negative reward $r_t < 0$ is obtained. However, a significant penalty is incurred if the patient dies after the treatment at. The reward at the time t preceding death is set to −15, i.e., $r_t = -15$.

In the current study, we shall employ Q-learning (QL) [25, 32], Deep Q-Network (DQN) [33-36], Double DQN (DDQN) [37, 38], Batch-Constrained deep Q-learning (BCQ) [39] and Conservative Q-Learning (CQL) [40] with rewards derived from the SOFA or cxSOFA scores to optimize the treatment strategy. Accordingly, SOFA-QL, SOFA-DQN, SOFA-DDQN, SOFA-BCQ, and SOFA-CQL denote the corresponding models when $r_t^{SOFA}$ is used, while cxSOFA-DQN, cxSOFA-DDQN, cxSOFA-BCQ, and cxSOFA-CQL refer to the models when $r_t^{cxSOFA}$ is used.

As we utilize RL to derive optimal strategy from historical databases such as MIMIC-IV and eICU, it seems more appropriate to employ offline methods like BCQ, CQL. In the medical realm, the actions permissible in a given state are constrained by medical guidelines. Similarly, the transition from a state-action pair $(s, a)$ to a new state $s'$ should not be generated randomly. To strictly adhere to these constraints, we not only conducted preliminary experiments with online reinforcement learning algorithms such as QL, DQN, and DDQN—treating each treatment trajectory as a sequence of continuous therapeutic interactions—but also systematically incorporated advanced offline reinforcement learning methods, namely BCQ and CQL. By comparing the performance of online and offline algorithms on identical medical trace data, this approach enables a comprehensive evaluation of the strategy optimization effectiveness of different models under clinical constraints.

### *C. Continuous Score Function cxSOFA*

The reward function in RL plays a critical role, and designing an appropriate reward function is a significant challenge [41]. The original SOFA score [5] has been used to design reward functions in previous studies [17, 23]. SOFA can be treated as a sum of six-step functions corresponding to respiratory, nervous, cardiovascular, liver, coagulation, and kidney systems, and it is widely used for the screening and evaluation of septic patients [3]. However, the discrete SOFA score is too crude as a basis for the reward function to accurately reflect changes in patient health before and after treatment. For instance, in the SOFA score's coagulation system assessment, a platelet count of 102 or 149 both results in a score of 1, while a count of 99 yields a score of 2. Since 102 is closer to 99 than to 149, its score should reflect proximity to 99. Yet, the actual scoring system shows a smaller difference between 102 and 149 than between 102 and 99, which is counterintuitive. To address this issue, we smooth the discrete SOFA score to a continuous score, cxSOFA: $S \mapsto \mathbb{R}$, by smoothing each step function, as defined below.

$$cxSOFA(x) = \sum_{1 \le i \le 6} f_i(x) \quad (3)$$

$$f_1(x) = \max\left(0, 4 - \frac{x_0}{100}, 4 - \frac{x_1}{100}\right)$$

$$f_2(x) = \max\left(0, -1.367 \cdot 10^{-6} \cdot x_2^3 + 4.075 \cdot 10^{-4} \cdot x_2^2 - 0.0573 \cdot x_2 + 4\right)$$

$$f_3(x) = \min\left(4, 1.137 \cdot 10^{-7} \cdot x_3^4 - 3.613 \cdot 10^{-5} \cdot x_3^3 + 2.685 \cdot 10^{-3} \cdot x_3^2 + 9.831 \cdot 10^{-3}\right),$$

$$f_4(x) = \max\left(\max\left(0, -\frac{x_4}{5} 14\right), \min(4, -3.365 \cdot 10^{-4} \cdot x_5^3 - 6.254 \cdot 10^{-5} \cdot x_5^2 + 0.208 \cdot x_5 + 2), 2bool(x_6, 0), \min(bool(x_7 > 0), 10 \cdot x_7 + 3), \min(bool(x_8 > 0), 10 \cdot x_8 + 3)\right)$$

$$f_5(x) = -1.404 \cdot 10^{-4} \cdot x_9^4 + 3.797 \cdot 10^{-3} \cdot x_9^3 - 4.175 \cdot 10^{-2} \cdot x_9^2 - 2.363 \cdot 10^{-2} \cdot x_9 + 4$$

$$f_6(x) = \max\left(\min(4, \max(0, 1.007 \cdot 10^{-9} \cdot x_{10}^4 - 48.889 \cdot 10^{-7} \cdot x_{10}^3 + 2.336 \cdot 10^{-4} \cdot x_{10}^2 - 7.181 \cdot 10^{-3} \cdot x_{10})), \min(4, \max(0, 2.015 \cdot 10^{-8} \cdot x_{11}^3 + 3.811 \cdot 10^{-6} \cdot x_{11}^2)) - 8.523 \cdot 10^{-3} \cdot x_{11} + 4.696\right)$$

where $x = (x_0, \dots, x_{11})$ with the entry $x_i$ for $0 \le i \le 11$ being the values of vital signs in Table I, and $f_j(x)$ defined as following equations (4)-(9) for $j = 1, 2, \dots, 6$ are smooth fittings to the corresponding step function through polynomial functions, $bool(x_i > 0)$ is the boolean predicate function, its value is 1 if $x_i > 0$ is true and otherwise is 0.

As shown in Fig.3, the same SOFA score can correspond to varying cxSOFA scores, indicating that a minor change in the SOFA score might result in a significant difference in the cxSOFA score. Consequently, the cxSOFA score, in comparison to the discrete SOFA score as state s, offers a more nuanced reflection of the physiological state of a patient.

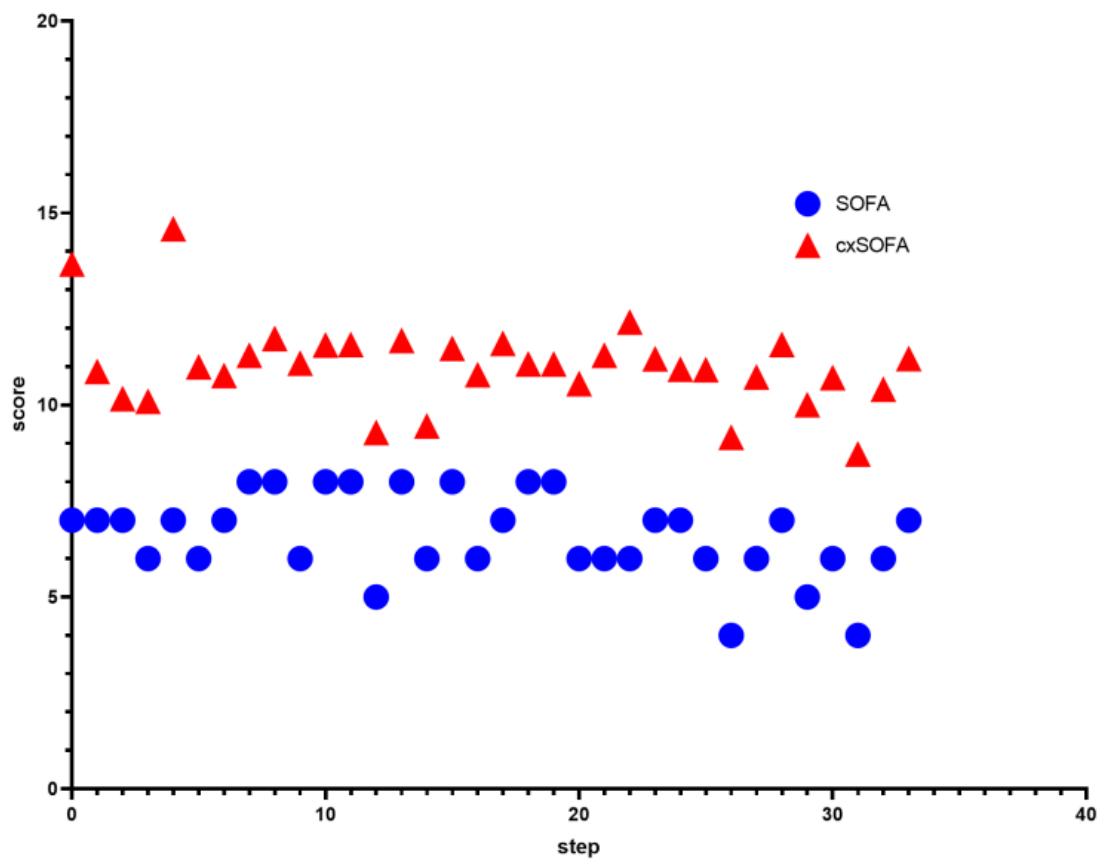

Fig. 3. Distribution of SOFA and cxSOFA scores throughout a patient's treatment trace.

### D. Data-Driven Concept of Good/Bad Treatment

In the training of continuous state spaces RL models DQN and DDQN, determining the optimal point to terminate the training process or establish convergence remains a significant challenge [42]. Substantially, it amounts to establishing strategy evaluation methods by which we can select the optimal strategy from ones trained with different epochs. A simple data-driven measurement was proposed[43], which divides the set P of the actual treatments of all physicians

$$ep = (s_0, a_0; s_1.a_1; \ldots; s_{n-1} = s_{ter}, a_{n-1} = \emptyset)$$

into two sets: $P_G$ and $P_B$, representing "good" and "bad" physicians' actual treatments, respectively. Specifically, $P_G$ contains all ep such that $s_{ter}$ is not death; and $P_B$ contains all ep such that $s_{ter}$ is death. This evaluation tried to assess the entire optimal strategy based on the actual patients' outcome, which would decrease the power and clinical significance of assessing every single decision-making during the whole process. As a result, it cannot evaluate the strategies on the data consisting of no death cases. To address this issue, we introduce novel notions of "good" and "bad" physicians' actual treatments. First, we define a stepwise "good" action if $r_t > 0$ and otherwise is a "bad" action. Based on these notions, we classify an episode into the class of "good" physicians' actual treatments if the proportion of "good" actions in all actions of the episode is beyond a given threshold τ, and otherwise into the class of "bad" physicians' actual treatments. Formally, let

$$N_g = |\{t: r_t \geq 0\}| \quad (4)$$

where $r_t$ is a reward function like $r_t^{SOFA}$ or $r_t^{cxSOAF}$, and the subscript g stands for "good". The effective rate $\rho_{ep}$ of this episode ep is defined as

$$\rho_{ep} = \frac{N_g}{n} \quad (5)$$

We define a threshold τ, referred to as the lowest effectiveness rate. For all episodes ep in the set P, if $\rho_{ep} \geq \tau$, we classify this physicians' actual treatment as a "good" physicians' actual treatment and add it to the set $P_G$. Otherwise, it is classified as a "bad" physicians' actual treatment and added to the set $P_B$.

## III. Novel Assessment Metrics

Assume $Q: (S, A) \mapsto \mathbb{R}$ is the value function learned by an RL method. Let $\pi^*$ and $\pi'$ denote the best and worst AI strategy, respectively, which are defined as

$$\pi^*(s) = \arg\max_{a \in A} Q(s,a) \quad (6)$$

$$\pi'(s) = \arg\min_{a \in A} Q(s,a) \quad (7)$$

for each state $s \in S$.

It is worth noting that the evaluation of $Q(s,a)$ based on merely the optimal strategy $\pi^*$ derived by $Q(s,a)$ may lose some information inside $Q(s,a)$. For a comprehensive evaluation of $Q(s,a)$, we suggest the consideration of the worst AI strategy $\pi'$ also derived by $Q(s,a)$. Within the set $P_G$ and the set $P_B$, we compute the optimal strategy $\pi^*(s_t)$ and the worst strategy $\pi'(s_t)$ for each state $s_t$ in each episode $ep = (s_0, a_0; s_1.a_1; \ldots; s_{n-1} = s_{ter}, a_{n-1} = \emptyset)$. Subsequently, the similarity between the actually treatment action $a_t$ and the optimal strategy $\pi^*(s_t)$, denoted as $sim_o(ep,\ t) = \frac{1}{1+0.25*|a_t - \pi^*(s_t)|}$, was calculated for every time step, as was the similarity to the action recommended by the worst strategy $\pi'$, denoted as $sim_w(ep,\ t) = \frac{1}{1+0.25*|a_t - \pi'(s_t)|}$.

Within the "good" physicians' actual treatments $P_G$: For each episode ep, we calculated:

$$N_{og}(ep) = |\{t: sim_o(ep,t) \geq sim_w(ep,t)\}| \quad (8)$$

$$\rho_{og} = \frac{N_{og}(ep)}{n} \quad (9)$$

$$\rho_{wg} = 1 - \rho_{og} \quad (10)$$

If $\rho_{og}$ exceeded a predefined threshold τ, the episode was classified into the opt-pol; otherwise, they were assigned to the wrt-pol. The average similarity $OG = \frac{\sum_{ep \in opt-pol} \frac{\sum_t sim_o(ep,t)}{|ep|}}{|opt-pol|} * \rho_{og}$ was then computed for episode in the opt-pol , and the average similarity $WG = \frac{\sum_{ep \in wrt-pol} \frac{\sum_t sim_w(ep,t)}{|ep|}}{|wrt-pol|} * \rho_{wg}$ was computed for those in the wrt-pol within $P_G$.

Within the "bad" physicians' actual treatments $P_B$: For each episode, we calculated:

$$N_{wb}(ep) = |\{t: sim_w(ep,t) \geq sim_o(ep,t)\}|, \quad (11)$$

$$\rho_{wb} = \frac{N_{wb}}{n} \quad (12)$$

$$\rho_{ob} = 1 - \rho_{wb} \quad (13)$$

If $\rho_{wb}$ exceeded a predefined threshold τ, the episode was classified into the wrt-pol; otherwise, they were assigned to the opt-pol. The average similarity $OB = \frac{\sum_{ep \in opt-pol} \frac{\sum_t sim_o(ep,t)}{|ep|}}{|opt-pol|} * \rho_{ob}$ was then computed for episode in the opt-pol, and the average similarity $WB = \frac{\sum_{ep \in wrt-pol} \frac{\sum_t sim_w(ep,t)}{|ep|}}{|wrt-pol|} * \rho_{wb}$ was computed for those in the wrt-pol within $P_B$.

Inspired by the confusion matrix, we compare the optimal/the worst AI strategy and "good"/"bad" physicians' actual treatments. Accordingly, we calculate the consistency/disagreement measures between the worst strategy and the "good" physicians' actual treatments, as well as between the worst strategy and the "bad" physicians' actual treatments, to provide a comprehensive assessment of the performance of the AI strategy. As a result, this yields the treatment effect comparing matrix as Table III. The consistency/disagreement metrics can be the action similarity

TABLE III
TREATMENT EFFECT COMPARISON MATRIX (TECM)

| Symbol | Optimal strategy: $\pi^*$ | Worst strategy: $\pi'$ |
|---|---|---|
| "good" actual treatments: $P_G$ | $OG$ | $WG$ |
| *"bad" actual treatments:* $P_B$ | $OB$ | $WB$ |

$P_G$, "good" physicians' actual treatments; $P_B$, "bad" physicians' actual treatments; OG, a consistency measure between the optimal strategy and the "good" physicians' actual treatments; OB, a disagreement measure between the optimal strategy and the "bad" physicians' actual treatments; WG, a disagreement measure between the worst strategy and the "good" physicians' actual treatments; WB, a consistency measure between the worst strategy and the "bad" physicians' actual treatments.

rate, relative gain, or other measuring methods.

For an AI strategy, higher consistency measures (between OG and WB) and lower disagreement measures (between OB and WG), higher O-gap $\Xi^{o} = \text{OG}-\text{OB}$ or W-gap $\Xi^{w} = \text{WB}-\text{WG}$ indicates better overall performance, assuming well-defined "good"/"bad" physicians' actual treatments and consistency/disagreement metrics.

However, the aforementioned six metrics would not work well in some circumstances. For example, consider two AI strategies with TECM shown in Table Ⅳ. In this table, GPAT, and BPAT denote "good" physicians' actual treatments and "bad" physicians' actual treatments, respectively; opt-pol and wrt-pol denote optimal strategy and worst strategy, respectively.

TABLE IV
COMPARISON OF TECM

| | $\pi_1$ | | $\pi_2$ | |
|---|---|---|---|---|
| | Opt-pol | Wrt-pol | Opt-pol | Wrt-pol |
| GPAT | $OG = 80\%$ | $WG = 30\%$ | $OG = 60\%$ | $WG = 50\%$ |
| BPAT | $OB = 40\%$ | $WB = 70\%$ | $OB = 20\%$ | $WB = 40\%$ |

GPAT, "good" physicians' actual treatments; BPAT, "bad" physicians' actual treatments; OG, OB, WG, and WB are defined as Table III.

AI strategy $\pi_1$ has a higher OG than $\pi_2$ but a lower WB, and its WG is lower than $\pi_2$ but its OB is higher. This makes it challenging to determine which strategy performs better overall. To address the issue, we propose two novel assessment metrics based on the TECM to evaluate the overall performance of AI strategies: comprehensive confidence and comprehensive bias. Inspired by the F1-score, we combine OG, OB, WG, and WB in the TECM into a single metric called comprehensive confidence $\sigma$ defined as

$$\sigma = \frac{2 \cdot \text{OG} \cdot \text{WB}}{\text{OG} + \text{WB}} \cdot \frac{\text{OG} + \text{WG}}{2 \cdot \text{OG} \cdot \text{WG}} \tag{14}$$

The higher the $\sigma$, the better the AI strategy.

Given a value function Q, $\text{OG} - \text{WG} > 0$ means that the optimal strategy is closer to the "good" actual treatments than the worst strategy, in contrast, $\text{WB} - \text{OB} > 0$ means that the worst strategy is closer to the "bad" actual treatments than the optimal strategy. Accordingly, $\text{OG} - \text{WG} > \text{WB} - \text{OB}$ indicates that the distance between the optimal strategy and "good" actual treatment is larger than that between the worst strategy and "bad" actual treatment. Intuitively, the optimal strategy may improve the treatment effect more than how the worst strategy worsens the treatment effect, making it more suitable for scenarios that require aggressive treatment strategies. In contrast, $\text{OG} - \text{WG} < \text{WB} - \text{OB}$ means that the worst strategy may worsen the treatment effect more than how the optimal strategy improves the treatment effect, making it more suitable for scenarios that require conservative treatment strategies. To capture these notions, we use the comprehensive bias $\mu$, defined as

$$\mu = (\text{OG} - \text{WG}) - (\text{WB} - \text{OB}) \tag{15}$$

to indicate whether the strategy is inclined to select aggressive or conservative treatment strategies. $\mu > 0$ indicates that the optimal AI strategy is better at finding the action that maximizes the treatment effect, i.e., aggressive treatment strategy. Conversely, $\mu < 0$ indicates that the optimal AI strategy tends to adopt conservative treatment strategies. In generally for two AI strategies $\pi_1$ and $\pi_2$, $\pi_1$ is said to be more conservative than $\pi_2$ if $\mu(\pi_1) \le \mu(\pi_2)$, where $\mu(\pi_k)$is the comprehensive bias of $\pi_k$ for $k = 1, 2$.

Since O-gap $\Xi^{o}$, W-gap $\Xi^{w}$, $\sigma$ and $\mu$ are derived from TECM, the class of those assessments including TECM is denoted by TECM*. By using the TECM*, we can comprehensively evaluate the performance of AI strategies using historical data on physicians' actual treatments.

## *A. Strategy Selection via TECM*-Principle*

After SOFA-QL, SOFA-DQN, cxSOFA-DQN, SOFA-DDQN, cxSOFA-DDQN, SOFA-BCQ, cxSOFA-BCQ, SOFA-CQL, and cxSOFA-CQL are trained, it is a key challenge to select the best strategy among them. Because an RL model may have over-estimation issues (e.g., Dueling DQN model [35]) or under-estimation issues (e.g., Dueling Double Deep Q-Network [44]) on $Q(s, a)$. It is not a good option to choose the strategy with the maximal return value by using $Q(s, a)$.

To address the issue above, we leverage TECM* to establish an AI strategy selecting principle as follows. Comparing the O-gaps $\Xi^{o}(\pi_i)$ and W-gaps $\Xi^{w}(\pi_i)$ of two strategies $\pi_i$ for $i = 1, 2$, the optimal strategy is determined by the following three rules:

- If $\Xi^{o}(\pi_i) \ge \Xi^{o}(\pi_j)$, i.e., the optimal $\pi_i^*$ is more close to a "good" physicians' actual treatment than $\pi_j^*$ , and $\Xi^{w}(\pi_i) \ge \Xi^{w}(\pi_j)$, i.e., the worst $\pi_i'$ is more close to "bad" physicians' actual treatment than $\pi_j'$, for $i \ne j \in \{1, 2\}$ then $\pi_i$ is said better than $\pi_j$.
- Otherwise $\Xi^{o}(\pi_i) \ge \Xi^{o}(\pi_j)$ and $\Xi^{w}(\pi_i) < \Xi^{w}(\pi_j)$ for $i \ne j \in \{1, 2\}$, i.e., $\pi_i$ and $\pi_j$ do not have the same trend to physicians' actual treatment from "good" and "bad" aspects, the optimal strategy is determined by comprehensive confidence $\sigma$ or comprehensive bias $\mu$. When $\sigma$ is used, $\pi_i$ is said better than $\pi_j$ if $\sigma(\pi_i) \ge \sigma(\pi_j)$.
- When $\mu$ is used in second case, assuming $\sigma(\pi_i) \ge \sigma(\pi_j)$, if a conservative strategy is chosen then $\pi_j$ is said better than $\pi_i$, otherwise $\pi_i$ is the better one.

For convenience, we call the above rules the TECM*-principle.

## *B. η-TECM* Termination Criterion*

In the training process of Q-type models, different training epochs usually output different strategies. A basic question is what the termination criterion for the training process should be. For RL models operating in continuous state spaces, such as DQN and DDQN, determining convergence is a nontrivial task[42]. Unlike their discrete counterparts, the vast and often infinite state space makes it challenging to establish definitive convergence criteria or conditions.

To address the issue of termination criterion, based on the TECM*-principle, we propose a data-driven quantitative termination condition for Q-type algorithms, named η-TECM* termination criterion as follows. Once the optimal strategy is not chosen by the previous three-rule principle on the η-highest epochs, the algorithm should terminate, where $\eta \geq 1$ is a hyperparameter.

## IV. Experiment

### A. Descriptive Characteristics

As shown in Fig.1, this study ultimately included a total of 5,896 patients for analysis: 3,879 patients from MIMIC-IV and 133 patients from eICU, who received abdominal surgery, along with 516 patients from MIMIC-IV and 1,368 patients from eICU, who did receive non-abdominal surgery. The descriptive baseline for abdominal surgery was shown in Table V.

### B. Comparison Results of SOFA and cxSOFA

To evaluate the performances of SOFA models (SOFA-QL, SOFA-DQN, SOFA-DDQN, SOFA-BCQ, SOFA-CQL) and cxSOFA models (cxSOFA-DQN, cxSOFA-DDQN, cxSOFA-BCQ, cxSOFA-CQL), respectively, we train them by using the same underlying data with the same batch size and epochs. We employed the comparison method [43] to evaluate the performances of those models on the validation test datasets with the same episodes. Accordingly, we checked whether the AI strategy could reduce the mortality rate or in-hospital stay by comparing episodes in the test datasets that followed the AI strategy with those that did not. For research justification, we compared the group that followed the AI strategy by matching the action in the episode. If the action of the physicians' actual treatment matched the AI strategy recommendation and its period comprised more than τ of the whole data period of the episode, we considered them as AI strategy followers. In the previous study [43], τ is set as 50%. In this study, we investigate the cases for various τ=50%, 60%, 65%, 70%, 75%, 80%, 90%, Fig.4 and Fig.5 reports the results for the moderate τ=70%. We computed the two groups' mortality rate and in-hospital stays concerning physicians' actual treatment and various AI strategies. The differences in mortality rate and in-hospital stay between different models were determined by

TABLE V
Baseline Information of Patients With UFH After Abdominal Surgery

| Indicators | MIMIC-IV | eICU |
|---|---|---|
| Age [years old, M (P25, P75)] | 68.67(57.72,80.39) | 69.16(57.42,80.75) |
| BMI [kg/m2, M (P25, P75)] | 28.21(24.38,33.71) | 27.16(23.26,32.53) |
| Female, n(%) | 280(50) | 964(51.6) |
| Number of diagnoses [n,M(P25,P75)] | 23.0(17.0,30.0) | 10.0(7.0,14.0) |
| Emergency, n(%) | 365(65.2) | 819(43.9) |
| Respiratory failure | 243(43.4) | 830(44.5) |
| Heart failure | 152(27.1) | 440(23.6) |
| Diabetes | 181(32.3) | 642(34.4) |
| COPD | 65(11.6) | 222(59.6) |
| Nephropathy | 367(65.5) | 1090(58.4) |
| Hepatopathy | 114(20.4) | 442(23.7) |
| Cancer | 124(22.1) | 386(20.7) |
| Hepatobiliary and pancreas | 171(30.5) | 116(6.2) |
| Female Reproductive | 3(0.5) | 0(0) |
| Lymphatic (spleen) | 2(0.4) | 0(0) |
| Urinary | 94(16.8) | 547(29.3) |
| Gastrointestinal | 290(51.8) | 1251(67.0) |
| 90-day mortality, n(%) | 126(22.5) | 273(14.7) |
| Length of stay [day,M(P25,P75)] | 13.10(7.04,24.64) | 11.58(5.92,15.36) |
| Blood product transfusion, n(%) | 266(47.5) | 20(1.1) |
| Coagulopathy, n(%) | 25(4.46) | 237(12.7) |
| Initial fluid resuscitation, n(%) | 411(73.4) | 1299(69.6) |
| IFRV [ml,M(P25,P75)] | 56.03(0,161.36) | 55.41(0,157.84) |
| Vasopressor, n(%) | 420(75.0) | 1075(57.6) |
| VMD [ml,M(P25,P75)] | 2.97(0.01,16.0) | 4(0,12.0) |

Data are expressed as number (proportion) and median (IQR [range]). BMI, body mass index; COPD, chronic obstructive pulmonary disease; IFRV, initial fluid resuscitation volume; VMD, vasopressor maximum dose.

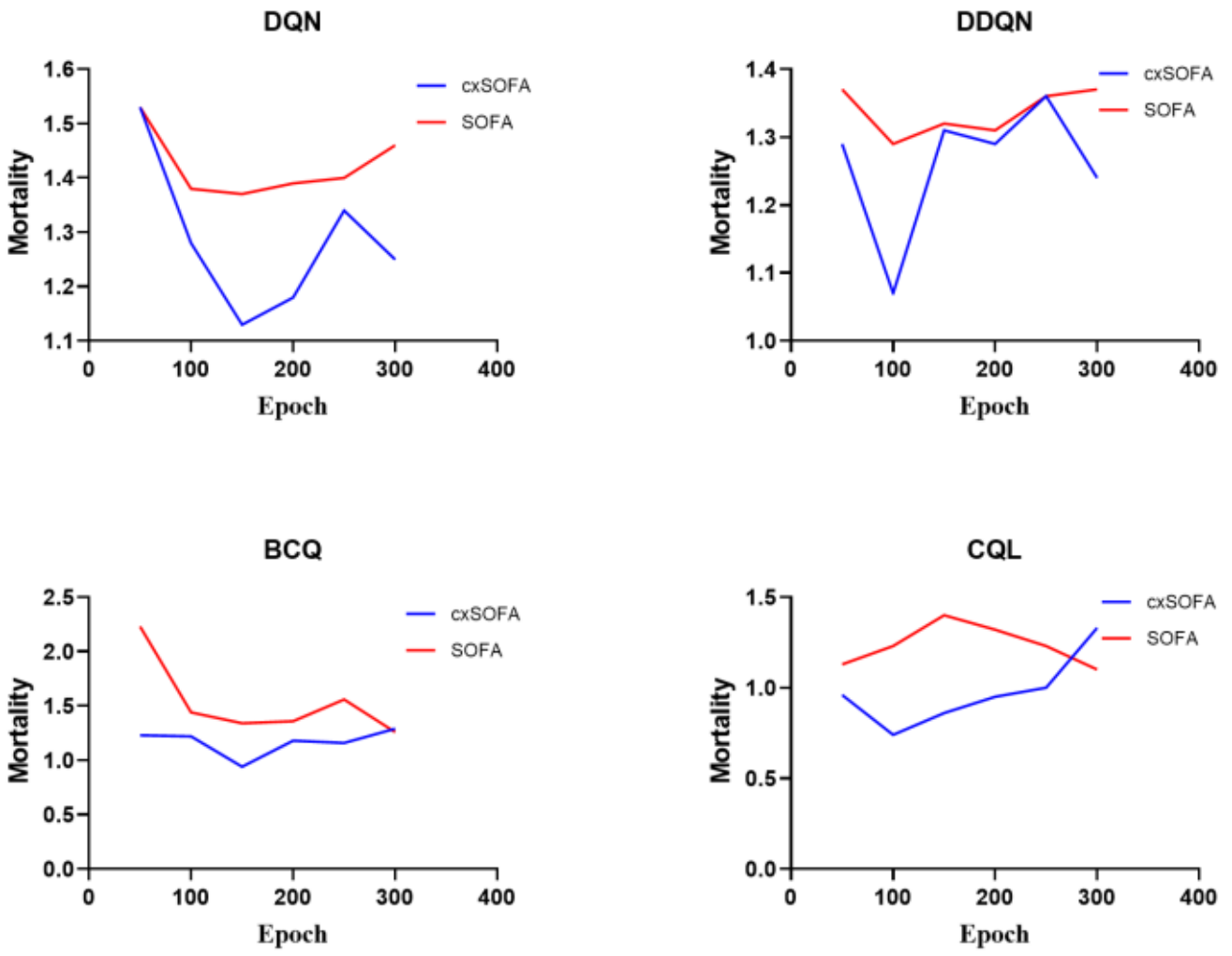

Fig. 4. Mortality rates with SOFA & cxSOFA

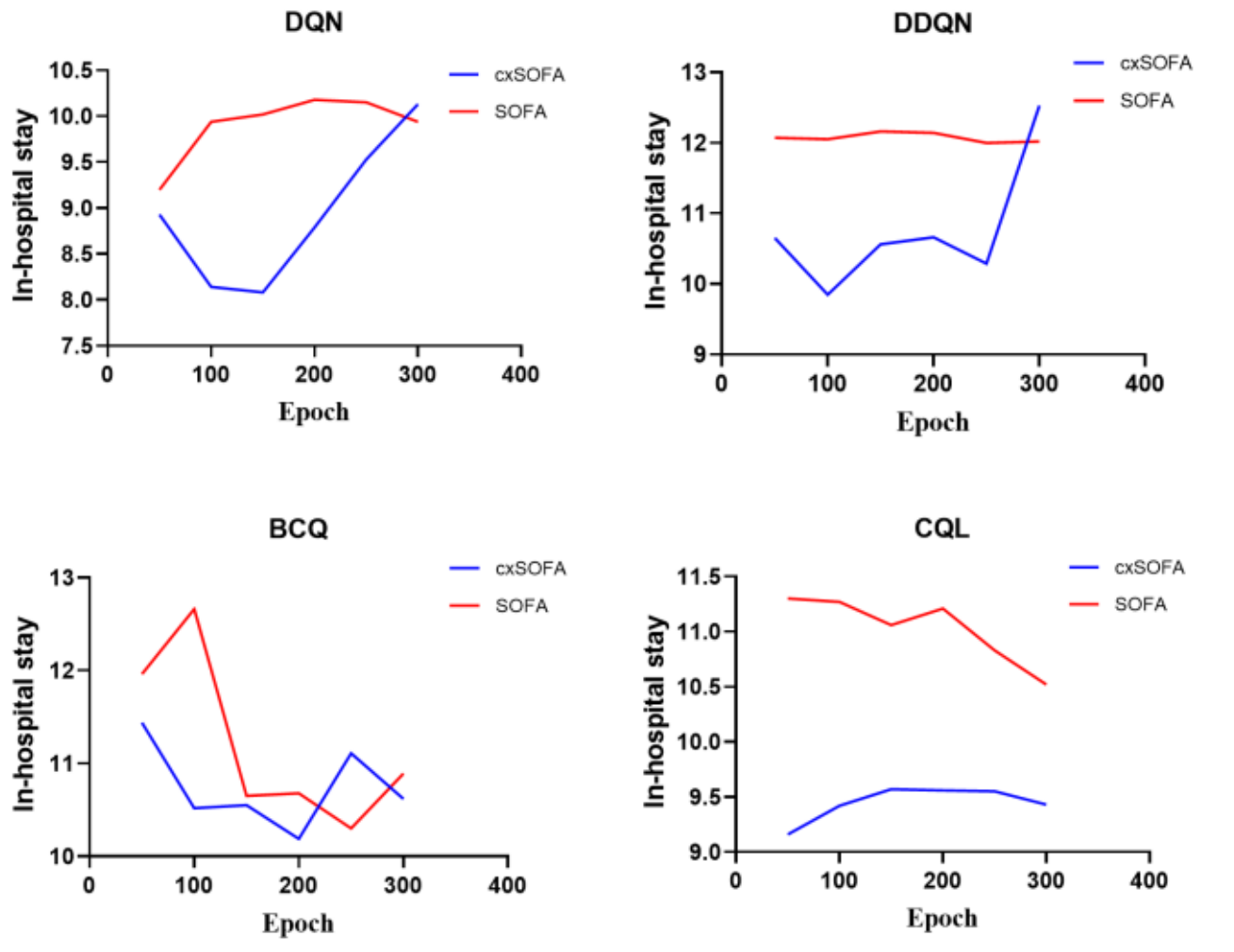

Fig. 5. Average in-hospital stays with SOFA & cxSOFA

the T-test. When the p-value is less than 0.05, the difference is statistically significant.

As shown in Fig.4 and Fig.5, optimized AI strategies based on cxSOFA as a reward function achieved better performance than those with SOFA, evidenced by lower mortality. Specifically, the average mortality in optimized strategy by DQN, DDQN, BCQ, CQL based on cxSOFA was statistically significantly lower than that based on SOFA (p-value=0.0489, 0.0467, 0.0254 and, 0.0125 respectively) Interestingly, comparing with the actual patients' outcomes in Table VI, the cxSOFA-version model showed a significant improvement in mortality (38.25%-59.56%) and in-hospital stay (5.04%-27.27%).

## C. *Usage of η-TECM* Termination Criterion*

Here we demonstrate how to apply η-TECM* termination

TABLE VI
MORTALITY RATES AND AVERAGE IN-HOSPITAL STAYS OF VARIOUS STRATEGIES

| PAT | | MoR | AIHS |
|---|---|---|---|
| | | 1.83% | 11.11 |
| QL | SOFA-QL | 1.35%, ↑26.23% | 10.45, ↑5.94% |
| DQN | SOFA-DQN | 1.37%, ↑25.14% | 10.18, ↑8.37% |
| | cxSOFA-DQN | 1.13%, ↑38.25% | 8.08, ↑27.27% |
| DDQN | SOFA-DDQN | 1.26%, ↑31.15% | 8.02, ↑27.81% |
| | cxSOFA-DDQN | 1.07%, ↑41.53% | 9.85, ↑11.34% |
| BCQ | SOFA-BCQ | 1.26%, ↑31.15% | 10.89, ↑1.98% |
| | cxSOFA-BCQ | 0.94%, ↑48.63% | 10.55, ↑5.04% |
| CQL | SOFA-CQL | 1.17%, ↑36.06% | 9.92, ↑10.71% |
| | cxSOFA-CQL | 0.74%, ↑59.56% | 9.42, ↑15.21% |

PAT, the physicians' actual treatment; MoR, mortality rate; AIHS, average in-hospital stay, ↑ indicates the improvement compared to the actual value in PAT , calculated as: $(\mathrm{MoR}_{\mathrm{PAT}} - \mathrm{MoR}_{\mathrm{AI}})/\mathrm{MoR}_{\mathrm{PAT}}$, or $(\mathrm{AiHS}_{\mathrm{PAT}} - \mathrm{AiHS}_{\mathrm{AI}})/\mathrm{AiHS}_{\mathrm{PAT}}$, where AI denotes SOFA-QL, SOFA-DQN, cxSOFA-DQN, SOFA-DDQN, cxSOFA-DDQN, SOFA-BCQ, cxSOFA-BCQ, SOFA-CQL, cxSOFA-CQL.

criterion to select one optimal strategy among several training results from various training epochs of the cxSOFA-CQL model. In this study, we set $\eta = 50$. The calculation of TECM* needs metrics between the AI's optimal/worst strategies and both "good"/"bad" physicians' actual treatments. We choose the action similarity rate in the previous study [17] as such metric in the demonstration. As described in the Methods section, we analyzed the AI strategies of cxSOFA-CQL for $\tau \in [0.5, 1]$. Fig.6 showed the TECM performance of different training epochs cxSOFA-CQL model on the validation datasets, and the relatively satisfied training epochs were [100, 200].

## D. *Application of TECM*-Principle*

In the performance evaluations of SOFA-QL, SOFA-DQN, cxSOFA-DQN, SOFA-DDQN, cxSOFA-DDQN, SOFA-BCQ, cxSOFA-BCQ, SOFA-CQL, and cxSOFA-CQL, we face the situation of picking the best strategy from several strategies generated by different AI algorithms. This can be done by the TECM*-principle assessment. This principle also can be applied to compare different strategies generated by various methods. For instance, $\pi_1$ and $\pi_2$ are AI strategies generated by QL and DQN. After these two methods obtained $\pi_1$ and $\pi_2$ by using their validation datasets, we employ the "good"/"bad" physicians' actual treatments derived from the testing datasets with the same underlying episodes to evaluate $\pi_1$ and $\pi_2$ by comparing them within $\tau \in [0.5, 1]$. To investigate the effectivness of cxSOFA, meanwhile comparing the AI strategies generated by DQN, DDQN, BCQ, and CQL, we implemented the assessments of AI strategies generated from SOFA-QL, SOFA-DQN, cxSOFA-DQN, SOFA-DDQN, cxSOFA-DDQN, SOFA-BCQ, cxSOFA-BCQ, SOFA-CQL, and cxSOFA-CQL, based on the validation datasets with the same underlying episodes.

We chose the action similarity rate [17] as the basic metric, and implemented the assessments of AI strategies generated from the previous nine models, based on the validation datasets with the same underlying episodes. For $\tau = 0.7$, the TECM* indicators based on action similarity rate (Fig.7), cxSOFA-CQL offers the best optimal AI strategies. Table VII reports the precise results w.r.t. Fig.7 for $\tau = 0.7$. All results in Table VII were subjected to the T-test with a p-value < 0.05, indicating them statistically significant. Among the four cxSOFA-models, cxSOFA-CQL demonstrates stable and balanced performance across the metrics of $\Xi^{o}$, $\Xi^{w}$, μ, and, σ. Furthermore, the optimized strategy by cxSOFA-CQL led to lower mortality and average in-hospital stay compared with that by the other three cxSOFA-models, and the differences were statistically significant. Therefore, for $\tau = 0.7$, the experiments recommend the AI strategy by cxSOFA-CQL. Our results also demonstrated that the termination criteria based on TECM* can properly select the optimal strategy.

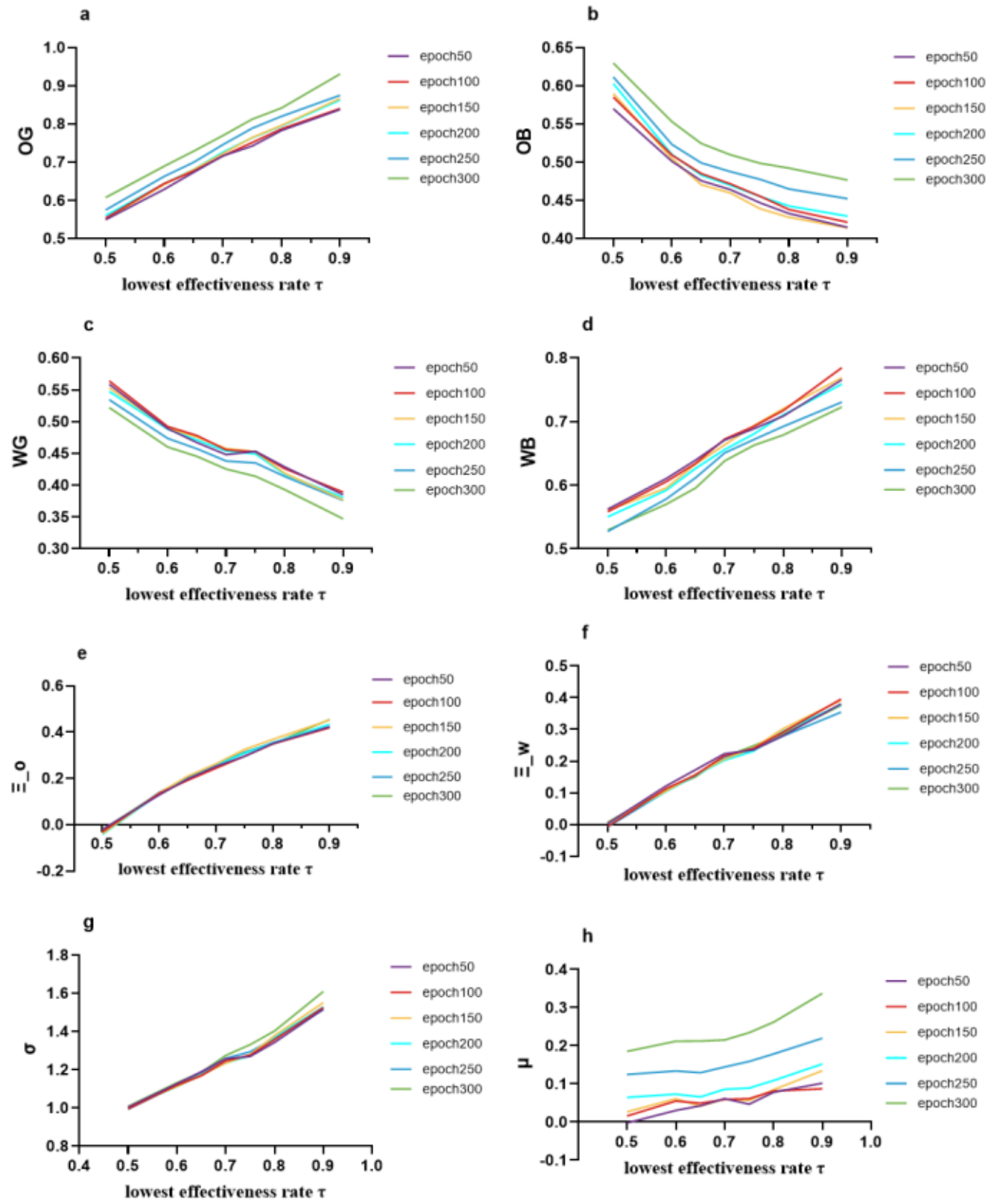

Fig. 6. TECM* of cxSOFA-CQL via action similarity rate. cxSOFA-CQL, CQL-Network with reward defined by cxSOFA; OG, OB, WG, and WB are defined as Table III; O-gap $\Xi^{o} = \mathrm{OG} - \mathrm{OB}$, W-gap $\Xi^{w} = \mathrm{WB} - \mathrm{WG}$; σ, comprehensive confidence, defined as (14); μ, comprehensive bias, defined as (15).

## V. Discussion

In this study, we proposed an RL-based framework to optimize personalized treatment strategies for surgical sepsis. First, we apply a cxSOFA score as the basis of reward function, to enhance the performance of Q-type models used in medical strategy optimization. Second, we initiate the study of the "worst" AI strategy in addition to the "optimal" AI strategy and introduce the concept of TECM in Table III. to assess treatment strategies by comparing "optimal" and "worst" AI treatment strategies with "good" and "bad" treatments, addressing the treatment selection issue outlined in the previous study [18]. Furthermore, we utilize TECM* to establish the η-TECM* termination criterion for the training process of reinforcement learning models concerned with treatment effects, thus tackling the challenge of defining termination conditions [24]. Finally, we apply the new framework to optimize the heparin treatment for surgical sepsis, across various Q-type models, revealing substantial improvements in treatment strategy and robustness across algorithms.

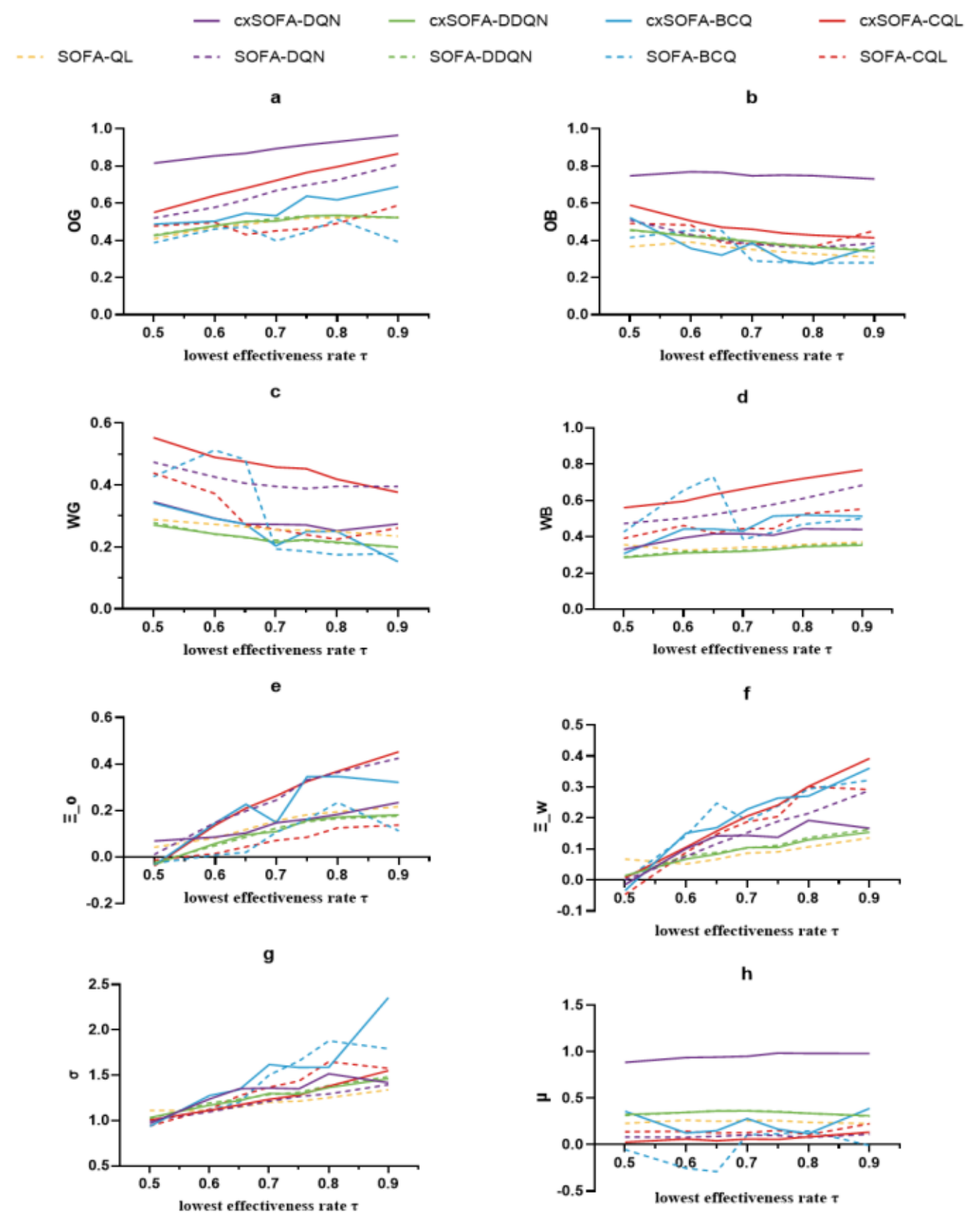

Fig. 7. TECM & comprehensive assessments. XX-YY, YY-type Q method with reward defined by XX, YY is one of QL, DQN, DDQN, BCQ, and CQL, XX is one of SOFA and cxSOFA; OG, OB, WG, and WB are defined as Table III; O-gap $\Xi^{o} = \mathrm{OG} - \mathrm{OB}$, W-gap $\Xi^{w} = \mathrm{WB} - \mathrm{WG}$; σ, comprehensive confidence, defined as (14); μ, comprehensive bias, defined as (15)

TABLE VII
TECM of Various Strategies

| | PAT | cxSOFA | | | |
|---|---|---|---|---|---|
| | | DQN | DDQN | BCQ | CQL |
| OG | ✗ | 0.5087 | 0.5222 | 0.689 | 0.7878 |
| OB | ✗ | 0.3953 | 0.3456 | 0.3678 | 0.3257 |
| WG | ✗ | 0.3838 | 0.1991 | 0.1525 | 0.0914 |
| WB | ✗ | 0.4845 | 0.3622 | 0.5129 | 0.3123 |
| $\sigma$ | ✗ | 1.3961 | 1.4837 | 2.3554 | 2.7315 |
| $\mu$ | ✗ | 0.1127 | 0.3064 | 0.3915 | 0.7097 |
| BE | ✗ | 150 | 100 | 200 | 150 |
| MoR | 1.83% | 1.13% | 1.07% | 0.94% | 0.74% |
| AIHS | 11.11 | 8.08 | 9.85 | 10.55 | 9.42 |

PAT, the physicians' actual treatment; BE, best epoch; QL, Q-learning; DQN, Deep Q-Network; DDQN, Double Deep Q-Network; BCQ, Batch-Constrained deep Q-learning; CQL, Conservative Q-Learning; OG, OB, WG, and WB are defined as Table III; σ, comprehensive confidence, defined as (14); μ, comprehensive bias, defined as (15); BE, best epoch, MoR, mortality rate; AIHS, average in-hospital stay.

Reinforcement learning (RL) [45], has been successfully applied in various medical domains, especially for chronic disease management, including HIV/AIDS [46], cancer [47], and diabetes [48]. RL aims to enhance decision-making through interactive experiences and evaluative feedback [24]. It addresses sequential decision-making problems involving sampling, evaluation, and delayed feedback, making it suitable for healthcare domains [49]. Unlike supervised [50] and unsupervised learning [51], RL does not require a correct label list. Unlike other traditional control-based methods, RL does not require a mathematically representative model of the environment [52]. RL formulates control strategies based on experience or historical data feedback, predicting states and rewards during learning [53]. However, applying RL to clinical situations presents a significant challenge to design appropriate reward functions [23, 24]. To address this issue, through polynomial fitting we extend the classical SOFA [5] to a continuous score cxSOFA aimed at reflecting nuanced variations in a patient's health state. As demonstrated in Fig.4, Fig.5 and Table VI, cxSOFA significantly enhances model performance compared to the discrete SOFA score [5]. We think that the outperformance of cxSOFA should come from its precise evaluation (as shown in Fig.3) of the healthy status of the patient.

Unlike domains such as gaming and robotics wherein AI strategies can be directly applied and immediate feedback obtained, medical scenarios prohibit direct experimentation on patients due to ethical, moral, and legal constraints [54]. The assessment of AI medical treatment strategy is also a great challenge [55], particularly in accurately evaluating the performance of AI strategies developed through RL [24]. To address the issue, we present data-driven quantitative criteria for the "good" and "bad" practical physicians' treatments based on the changes in SOFA/cxSOFA scores among states of consecutive times. The new notions of "good" and "bad" have two highlight features. First, it is different from classical definitions such as the previous studies [17, 23], which not only consider the final reward of the treatment strategy but also include the intermediate rewards. Second, the intermediate rewards are computed by a dynamic approach, but not restricted to a constant table like in a study mentioned before [19]. Importantly, these new concepts are grounded in data-driven, quantitative metrics, providing an objective assessment of the treatment strategy when compared to conventional expert evaluations [56]. Therefore, these features give a comprehensive assessment of the treatment strategy and bring hope for new therapy discovery.

In the process of AI into medical practice, there are various challenges including ethical considerations, biases, accountability, and regulation [54], with treatment selection being a particularly significant one [18]. The conventional choice of treatment strategy [17, 19, 23] relies on the $Q(s,a)$-function mostly shaped by the optimal actions. Like a coin has two faces, the $Q(s,a)$-function can give not only the optimal strategy but also the "worst" one. Therefore, in this study, we also initiate the investigation of the "worst" AI strategy to get some insight into $Q(s,a)$ and assist the treatment selection. Moreover, inspired by the notion of confusion matrix [57], we introduce the TECM based on the new definitions of "good" and "bad" treatments and notions of the "optimal" and "worst" AI strategy, to analyze the rationality of the $Q(s,a)$-function and so the optimal strategy for a comprehensive assessment. TECM provides a deep insight into AI strategy derived from $Q(s,a)$-function. We use TECM to establish a data-driven quantitative TECM*-principle for strategy selection in medical RL prior to randomized controlled trials (RCTs) in clinical testing. To the best of our knowledge, TECM*-principle is the first such data-driven criteria to address the treatment selection issue [18].

In particular, we employ TECM* to give the $\eta$-TECM* termination criterion for the training process of RL models concerned with the treatment effect, to address the challenge of termination conditions [24]. This novel evaluation framework allows for a more comprehensive assessment of AI strategies in healthcare settings, where direct experimentation is not feasible. By providing multiple perspectives on strategy performance, our approach enables researchers and clinicians to and implementing AI-driven treatment strategies. The $\eta$-TECM* termination criterion offers several advantages: i) Objective termination: It provides a data-driven, quantitative approach to determine when to stop training, reducing subjective decision-making. ii) Efficiency: By avoiding unnecessary training iterations, it can save computational resources and time. iii) Performance optimization: The three-rule principle ensures that the selected strategy is the best performer among those generated during training. iv) Applicability to healthcare: This approach addresses a key challenge in applying RL models to healthcare decision support systems. Fig.6, Fig.7, Table VI, and Table VII demonstrate that the $\eta$-TECM* termination criterion can successfully identify the optimal AI strategy from those generated by six Q-type models.

We have successfully demonstrated the effectiveness and application potential of the cxSOFA score and the novel assessment metrics TECM* for optimizing heparin management in postoperative sepsis treatment episodes. The comprehensive study involved the development, training, and testing of six distinct models using two publicly available datasets: eICU and MIMIC-IV, ensuring the reproducibility and robustness of our findings. The results, as illustrated in Fig.7, clearly indicate that the performance of AI strategies trained using the cxSOFA-CQL model is significantly superior to all other approaches. Key insights from this study include i) Superiority of cxSOFA; The continuous nature of cxSOFA allows for a more nuanced representation of patient healthy status, leading to improved strategy performance. ii) Effectiveness of CQL: Among the tested algorithms, CQL consistently outperformed other approaches when combined with cxSOFA; iii) Robustness across datasets and basic metrics: The superior performance of cxSOFA-CQL was consistently demonstrated in both the testing eICU and MIMIC-IV datasets, encompassing postoperative sepsis patients who did and did not receive heparin treatment after onset.

These findings have significant implications for AI-driven healthcare decision support, particularly in managing complex conditions like postoperative sepsis. By leveraging the cxSOFA score and advanced RL techniques, clinicians may be able to make more informed and personalized decisions regarding heparin administration, potentially leading to improved patient outcomes. Under the framework presented in Fig.1, several valuable avenues for further exploration emerge.

- Continuous Action Spaces: While the current study employs discrete action spaces for all six models, even those with continuous states, this approach may have limitations in certain medical scenarios. For instance, in optimizing heparin dosage, the AI strategy outputs an interval, which could potentially confuse users, especially given that action a4 in Table II represents a semi-interval. Strategy-based algorithms operating in continuous state and action spaces may yield superior performance in such scenarios, providing more precise and nuanced recommendations for heparin dosage management.
- Integration of Data & Expert Driven: Definition of "good" Physician Treatment: The current definition of a "good" physicians' actual treatment is data-driven. An alternative approach would involve medical experts selecting the episodes that constitute "good" physician treatment. This method would allow for the integration of expert knowledge into the models, potentially enhancing their clinical relevance and acceptability.
- Clinical Validation and Generalization: Although the model demonstrates satisfactory generalization in terms of TECM*, it is crucial to acknowledge that extensive multi-center prospective evaluations using real-time data and decision-making in clinical trials are necessary. These evaluations will help to address the challenges posed by the high heterogeneity of sepsis and the diverse healthcare settings in which the model may be applied. Such rigorous clinical validation is essential for ensuring the model's effectiveness and safety in real-world scenarios.

## VI. Conclusion

This study developed a new continuous score cxSOFA for more nuanced reward functions in RL methods and introduced a novel data-driven assessment metric TECM* for objectively and efficiently evaluating the treatment strategy. In particular, we presented a termination criteria for the training process of RL models. The continuous score cxSOFA and assessment metric TECM* are verified across multiple RL models and databases, demonstrating the effectiveness and potential for optimizing heparin management in postoperative sepsis treatment, suggesting broad applicability.